\documentclass[letterpaper,10pt,conference]{ieeeconf}

\usepackage{times}
\usepackage{graphicx}
\usepackage{amsmath,amssymb,amsopn,amstext,amsfonts}
\usepackage{cancel}
\usepackage[space]{cite}
\usepackage{pdfsync}
\usepackage{balance}
\usepackage{color}
\usepackage{mathtools}
\usepackage{algpseudocode}
\usepackage{algorithm}
\usepackage{bm}

\usepackage{diagbox}
\usepackage{float}
\usepackage{epstopdf}
\usepackage{pifont}
\usepackage{amsmath}
\usepackage{multirow}
\usepackage{url}
\usepackage{verbatim}
\usepackage[linkcolor=black,citecolor=black,urlcolor=black,colorlinks=TRUE]{hyperref}
\usepackage{booktabs}
\usepackage{graphicx}
\usepackage{subcaption}
\usepackage{stfloats}
\newcommand{\tabincell}[2]{\begin{tabular}{@{}#1@{}}#2\end{tabular}}

\graphicspath{{./}}
\DeclareGraphicsExtensions{.png,.jpg,.eps,.pdf}
\IEEEoverridecommandlockouts
\begin{document}
        \title{\LARGE \bf Dynamic Free-Space Roadmap for Safe Quadrotor Motion Planning}

\author{ Junlong Guo, Zhiren Xun, Shuang Geng, Yi Lin\textsuperscript{1}, Chao Xu, and Fei Gao
    \thanks{All authors except Yi Lin\textsuperscript{1} are with the State Key Laboratory of Industrial Control Technology, Zhejiang University, Hangzhou 310027, China, and also with the Huzhou Institute of Zhejiang University, HuZhou 313000, China. Corresponding author: Fei Gao.\ \textsuperscript{1}DJI, Shenzhen 510810, China. This work is supported by DJI.
    E-mails:{ \tt\small \{22060478, cxu, fgaoaa\}@zju.edu.cn}}
    }

    \maketitle
    \thispagestyle{empty}
    \pagestyle{empty}

		\begin{abstract}
        Free-space-oriented roadmaps typically generate a series of convex geometric primitives, which constitute the safe region for motion planning. However, a static environment is assumed for this kind of roadmap. This assumption makes it unable to deal with dynamic obstacles and limits its applications. In this paper, we present a dynamic free-space roadmap, which provides feasible spaces and a navigation graph for safe quadrotor motion planning. Our roadmap is constructed by continuously seeding and extracting free regions in the environment. In order to adapt our map to environments with dynamic obstacles, we incrementally decompose the polyhedra intersecting with obstacles into obstacle-free regions, while the graph is also updated by our well-designed mechanism. Extensive simulations and real-world experiments demonstrate that our method is practically applicable and efficient.
        \end{abstract}

	\section{Introduction}
	\label{sec:introduction}

    Mapping modules are of vital importance for safe motion planning. They often take raw sensor data as input then output abstract environment representation. However, voxel-based maps need to store cumbersome environment data, which is both memory and time-consuming. The resolution also limits it to represent the environment in a finer resolution \cite{roth1989building,hornung2013octomap}. Point cloud maps do not directly support dynamic environments or efficient queries for motion planning. Therefore, it is necessary for a map representation to be able to consume less memory, handle dynamic obstacles, and be convenient for motion planning.

    Free spaces, instead of obstacles, is useful information for robot navigation \cite{han2019fiesta}. Polyhedron-shaped free-space roadmap is a natural idea that uses a set of polyhedra to represent the union of free space of the environment. It has the following advantages: \textbf{1)} Polyhedron could tightly approximate free configurations and naturally suits for non-convex and unstructured environments. \textbf{2)} Connectivity between each polyhedron represents topological structures of environments that have significant importance in motion planning. \textbf{3)} Storage requirement is significantly lower because polyhedrons can represent vast space by few parameters. \textbf{4)} Analytical expression of the polyhedron is convenient to encode free space information for trajectory optimization and can easily add safety requirements.

        \begin{figure}[t]
		\centering
		\includegraphics[width=1.0\columnwidth]{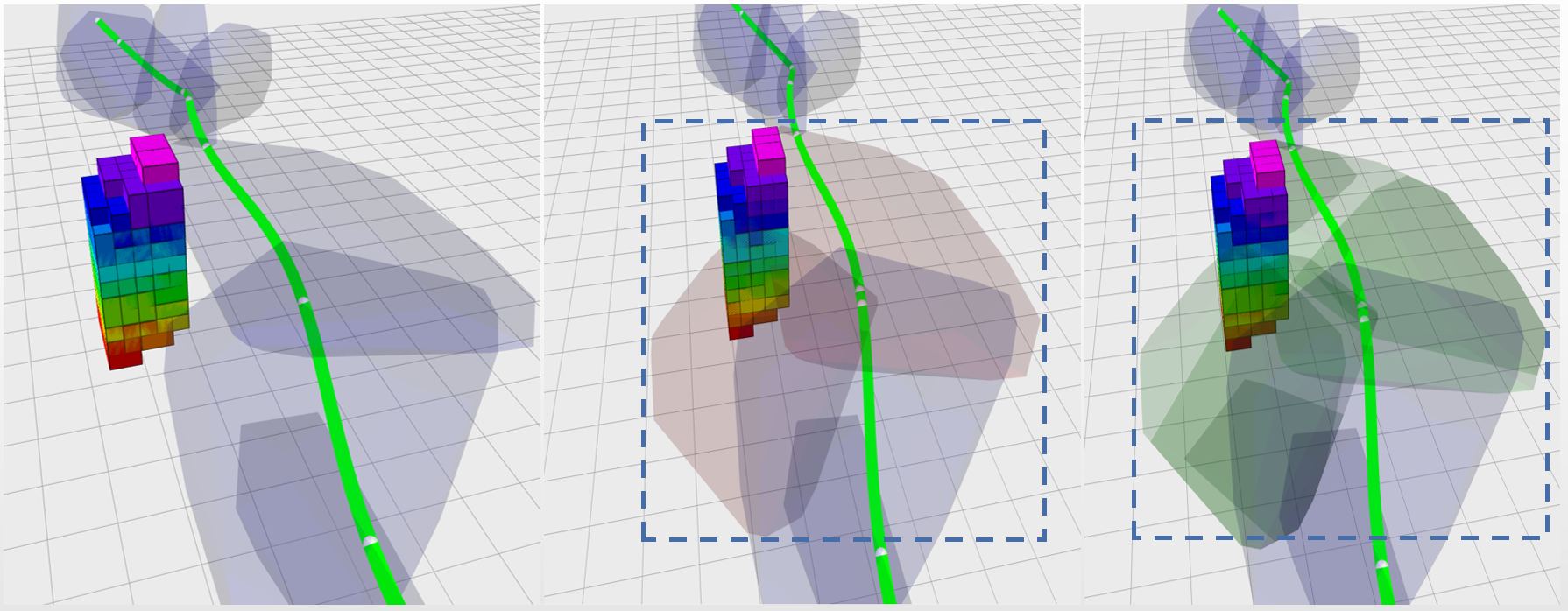}
		\caption{Illustration of dynamic free-space roadmap. The polyhedron colored with red overlap dynamic obstacles. Green polyhedra are obtained by decomposing the red one.}
	    \end{figure}

Although polyhedron-shaped free-space roadmap has the above advantages, existing methods cannot deal with dynamic environments,  which are common scenes for aerial robots. To our best knowledge, existing polyhedron-shaped free-space roadmaps have no efficient update strategy to deal with topological changes due to moving obstacles. Related works \cite{blochliger2018topomap,chen2019creating,deits2015efficient,wang2021robust,gao2020teach} all assume that the environment is static. To bridge this gap, this work introduces an efficient method that incrementally updates the free-space roadmap when there are moving obstacles on the map. Firstly, we use Iterative Regional Inflation by Semidefinite (IRIS) \cite{deits2015computing} to extract all free spaces in the map to generate polyhedra. Then a navigation graph will be built by checking connectivity between each polyhedron. After that, we build oriented bounding boxing\cite{2000Collision} for every moving obstacle, thus the infeasible region is approximated tightly. Then we utilize the hyperplanes of the bounding boxes to decompose polyhedra overlapping the obstacles into some small free polyhedra. Finally, we take advantage of the prior information of the graph to update the graph efficiently.

	    Contributions of this paper are:
	\begin{itemize}
		\item [1)]
		An efficient method is proposed that incrementally updates polyhedral maps in the dynamic environments yet preserves connectivity.
		\item [2)]
		An incremental graph updating mechanism is designed.
		\item [3)]
        An autonomous navigation system that uses the above methods is tested in simulation and real-world to validate that our methods are practical.
	\end{itemize}

    \section{Related Work}
	\label{sec:related_work}
	\subsection{Free-Space Oriented Map Representation}
	There are many works focusing on polyhedronizing free regions in environments for efficient navigation and trajectory optimization. Deits and Tedrake \cite{deits2015efficient} adopt IRIS which proposed by themselves to segment space into convex regions, then perform a mixed-integer optimization to obtain a collision free trajectory.
	Wang et al.\cite{wang2021robust} tightly approximates all free configurations by unions of polyhedra for encoding free space information into multi-aerial robots trajectory optimization. The work mostly relevant to ours is \cite{blochliger2018topomap}. The author extracts free spaces from noisy and partly incomplete visual SLAM data by voxel clustering and merging, then a topological navigable graph is constructed, which make global path planning easy and computationally inexpensive. In \cite{chen2019creating}, the author also use IRIS to generate polyhedra to represent obstacle-free regions from raw point cloud data, then projecting original sparse and noisy points onto the surfaces of polyhedra to form a dense version of point cloud which provides abundant environment information for human users. This is a rather inspiring work which for the first time combines point regulation and space convexification to create dense point cloud and navigable space.

	Nevertheless, these works still assume a static environment. In applications, there are more or less dynamic objects in the considered navigation scene.

	\subsection{Representation of Dynamic Environments}

    \setlength{\textfloatsep}{0.1cm}
    \begin{algorithm}[t]
    \caption{MapPolyhedronization}
    \label{algorithm:mappoly}
    \begin{algorithmic}[1]
        \State \textbf{Notation:}sampling time:$s$,
        maximum sampling times:$S$,
        expected ratio of $vol(\mathcal{P}_g)$ to $vol(\mathcal{F})$:$\rho_e$, $vol(\cdot)$ means volume.
        \State \textbf{Input:}$\mathcal{M}$,$S$,$\rho_{e}$,
        \State\textbf{Output:}$\mathcal{P}_g$
        \State   $\mathcal{F} \gets GetFreeRegion(\mathcal{M})$
        \State   $\mathcal{O} \gets GetObstacleRegion(\mathcal{M})$
        \State   $\mathcal{P}_g \gets \varnothing,
        \rho \gets 0, i \gets 0, s \gets 0$
        \While {$\rho < \rho_{e}$ and $s < S $}
        \State $q \gets random(\mathcal{F},\mathcal{P}_g)$
        \State $s \gets s+1$
        \State $bbox \gets BoundingBox(q)$
        \State $P_i \gets IRIS(q, bbox,\mathcal{M})$
        \If {$IsGoodPoly(P_i)$}
        \State $i \gets i+1$
        \State $\mathcal{P}_g.\gets \mathcal{P}_g \cup P_i$
        \State $\rho \gets UpdateRatio(\mathcal{P}_g,\mathcal{F})$
        \EndIf
        \EndWhile
        \State\Return  $\mathcal{P}_g$
    \end{algorithmic}
\end{algorithm}
\setlength{\floatsep}{0.1cm}

	Dealing with dynamic environments is a challenging topic in autonomous navigation. There are some approaches which handle it at a low level of map abstraction. In \cite{ajeil2020grid}, the author uses a probabilistic grid map to represent the environment and a hidden Markov model to occupancy state and state transition probabilities of the grid. The occupancy state of each grid will be updated when observations become available. Some other approaches explicitly model the dynamic objects instead. For example Anguelov \cite{anguelov2012learning} and Biswas \cite{biswas2002towards} compute the shape of dynamic models, use expectation maximization to identify dynamic parts of the maps that are created at different timestamps. In~\cite{wolf2005mobile}, two probability occupancy grid map are maintained accordingly for static parts and dynamic parts of the environment, these two grid maps will be updated by Bayesian rules once observations come from sensors. Lau et al.~\cite{lau2013efficient} develop algorithms that incrementally update grids which affected by changes. Meanwhile, Euclidean distance maps and generalized Voronoi diagrams can also be incrementally updated by their algorithms. Some other methods explicitly model aspects of the environment dynamics. For example, Wang et al. \cite{wang2021autonomous} and  Eppenberger et al. \cite{eppenberger2020leveraging} identify dynamic objects from environment based on the fact that the positions of dynamic objects change overtime. However these methods focus on tracking dynamic objects rather then represents the environments' dynamic parts.

    \section{METHODOLOGY}
	\label{sec:Method}
	\subsection{Map Polyhedronization}
	\label{sec:Map Polyhedronization}

	We combine IRIS\cite{deits2015computing} and RILS\cite{liu2017planning} to segment the environment into convex regions. IRIS proposed by Deits and Tedrake alternates two convex optimizations to generate a polyhedron by a given query seed: First a Quadratic Program (QP) generates a convex region represented by a set of hyperplanes. Second a Semidefinite Program (SDP) finds a maximum volume inscribe ellipsoid (MVIE) in the convex region that constructed in first step. RILS is similar to a non-iterative version of IRIS but more efficient because of a bounding box is applied to reduce obstacles need to considered. Like RILS does, we also use bounding box in IRIS to relieve computational burden.

	Let $\mathcal{M}$ denote the global occupied grid map. Note that the map $\mathcal{M}$ will only be used for map polyhedronization and not needed in subsequently motion planning. We denote by $\mathcal{F} \subseteq \mathcal{M}$ the free configurations in map $\mathcal{M}$ and by $\mathcal{P}_g$ the union of polyhedra where $\mathcal{P}_g \subseteq \mathcal{F}$ and
	    \begin{equation}
            \mathcal{P}_g = \bigcup_{i=0}^{N}P_i             ,
        \end{equation}

        \begin{equation}
            P_i = \{ x \in \mathbb{R}^{3} | A_ix \preceq b_i\}             ,
        \end{equation}

    The map polyhedronization algorithm is detailed as Algorithm \ref{algorithm:mappoly}. Firstly, we randomly sample a query point $\mathbf{q}=[x_p,y_p,z_p] \in \mathbb{R}^{3}$ that $\mathcal{F}$ while outside $\mathcal{P}_g$. Then IRIS takes the query point and the bounding box around it as input to generate a polyhedron $P_i$ in H-representation from $\mathcal{M}$. Each elements in $\mathcal{P}_g$ is managed in a hierarchical structure. In order to approximates the map more tightly, Algorithm \ref{algorithm:mappoly} will not stop until the ratio $\rho$ of $\mathcal{P}_g$ to $\mathcal{F}$ or sampling times $s$ over expected thresholds.

	Note that the purpose of polyhedronizing a map is to obtain the set of polyhedra which approximates all free spaces, thereby any other algorithms that generate a polyhedron by a query point can replace IRIS in Algorithm \ref{algorithm:mappoly}.

    \subsection{Polyhedron Decomposition and Restoration}
	\label{sec:Polyhedron Spliting and Recovery}
	\subsubsection{Preliminaries}
	 After map polyhedronization, we need to consider how to deal with the situation where obstacles overlap the polyhedra. Indeed the first thing is to distinguish point clouds which are inside the polyhedra. Therefore axis-aligned bounding box of these polyhedra are constructed, and managed by a multi-level segment tree $SegTree$ which is used for efficient stabbing queries \cite{MD2008Computational}. For a given point $\mathbf{p}=[x_p,y_p,z_p] \in \mathbb{R}^{3}$, the query time complexity is $O(log^3n+k)$ and $k$ bounding boxes enclose the point $\mathbf{p}$ can be obtained at the same time. We denote by $\mathcal{P}_l$ the polyhedra obtained by querying the $SegTree$ and enclosing the query point $\mathbf{p}$.

	 		\begin{algorithm}[t]
        	\caption{PolyhedronDecomposition}
        	\label{algorithm:polydec}
        	\begin{algorithmic}[1]
        	\State \textbf{Notation:}$\mathcal{P} = \{P_1,P_2,\dots,P_m\}$,
        	                    $P_i \in \mathcal{P}_g$
        	\State \textbf{Input:}$\mathcal{P}$,$obb$
        	\Function{PolyhedronDecomposition}{$\mathcal{P}$,$obb$ }
                \For {$P_i$ in $\mathcal{P}$}
                    \If{$P_i$.state == \textbf{complete}} %
                        \State $Decomposition(P_i,obb)$
                    \ElsIf{$P_i$.state == \textbf{decomposed}}
                        \State $\mathcal{P}_i \gets ObstainCollisionsons(P_i)$
                        \State \Call{PolyhedronDecomposition}{$\mathcal{P}_i$,$obb$}
                    \EndIf
                \EndFor
            \EndFunction

            \State

            \State \textbf{Input:}$P$,$obb$
            \Function{Decomposition}{$P$,$obb$}
            \State $P$.state $\gets$ \textbf{decomposed}
            \State $\mathcal{P}_t \gets \varnothing$
            \For{$H_i$ in $obb$}
                \State $P_i \gets Intersection(P,H_k)$
                \If{$IsGoodPoly(P_i)$ and $DeRedun(P_i,\mathcal{P}_t)$}
                    \State $\mathcal{P}_t \gets \mathcal{P}_t \cup P_i$
                    \State $P$.addson($P_i$)
                \EndIf
            \EndFor
            \EndFunction
           \end{algorithmic}
        \end{algorithm}

	 In order to separate obstacles from polyhedra, an oriented bounding boxe $obb_i$ are generated for each obstacle $O_i$, where
	     	    \begin{equation}\label{equ:obb}
                obb_i = \{ x \in \mathbb{R}^{3} | A_ix \preceq b_i\}.
            \end{equation}
     Any polyhedron intersecting the obstacle $O_i$ is denoted by $\mathcal{P}_i$. Similar to dilating obstacles in grid map, we expand the oriented bounding box by radius $r$ of the aerial robot to maintaining a configuration space.

	 \subsubsection{Polyhedron Decomposition and Restoration}
	 Benefiting from the convex sets property: \textit{the intersection of any collection of convex sets is convex}. It is obvious that any polyhedron determined by halfspace intersection is also a convex set. In viewing of this, a natural idea is to utilize hyperplanes of the bounding box to decompose the polyhedron to exclude the obstacle enclosed by oriented bounding box. The pipeline for Polyhedron Decomposition algorithm is detailed in Algorithm \ref{algorithm:polydec}.

\begin{figure}[t]
    \centering
    \includegraphics[width=1\columnwidth]{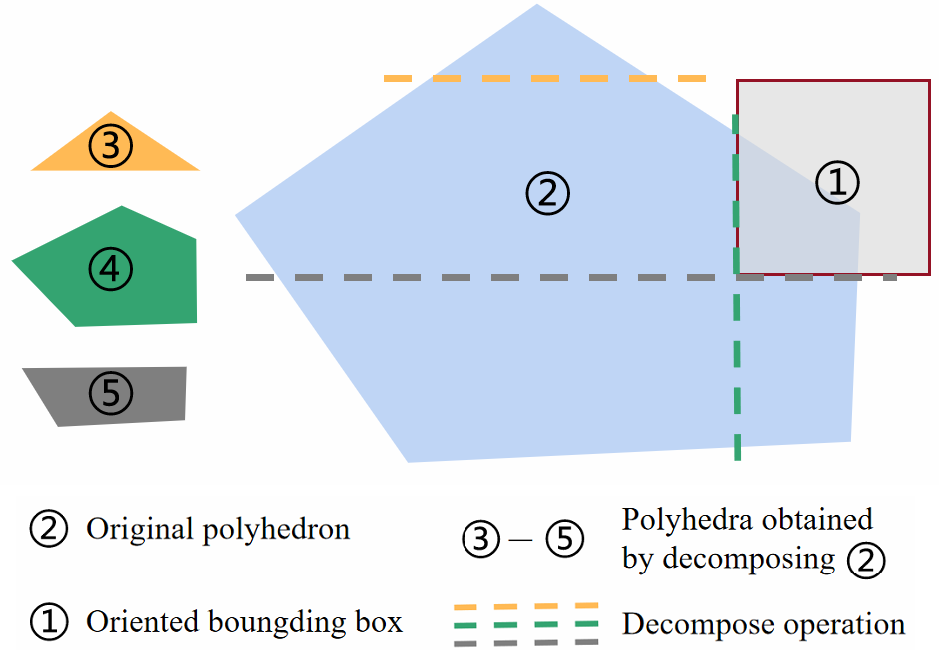}
    \caption{Polyhedron \ding{173} decomposed into polyhedra \ding{174}-\ding{176}  by oriented bounding box \ding{172}, however polyhedron \ding{174} is redundant since it insides \ding{175}.
    }
    \label{pic:Redundant}
\end{figure}

	 We use label \textbf{complete} and \textbf{decomposed} to distinguish whether a polyhedron is a complete one or a decomposed one. For every polyhedron in $\mathcal{P}$ that will be decomposed by $obb$, we first check the label of the polyhedron. Polyhedron that labeled as \textbf{complete} will be decomposed by each hyperplanes of $obb$. As illustrated in Fig. \ref{pic:Redundant},  redundant polyhedra in decomposition procedure will be abandoned. While for \textbf{decomposed} polyhedron, all elements in the next level of the hierarchy that intersecting with $obb$ will be decomposed again.

\setlength{\textfloatsep}{0.5cm}
\begin{algorithm}[b]
    \caption{PolyhedronRestoration}
    \label{algorithm:polyrestoration}
    \begin{algorithmic}[1]
        \State \textbf{Notation:}
        $obb_j \cap obb_i = \varnothing$,
        and $p_{ij}$ will be treated as a vector as input to Algorithms \ref{algorithm:polydec} for simplicity.
        \State \textbf{Input:}$O_i$,$obb_i$,$\mathcal{P}_i$
        \For{$p_{ij}$ in $\mathcal{P}_i$}
        \State $Restroation(p_{ij})$
        \For{\textbf{each} $obb_j$ that intetsecting with $p_{ij}$}
        \State $PolyhedronDecomposition(p_{ij},obb_j)$
        \EndFor
        \EndFor
        \State\Return  $\mathcal{P}_g$
    \end{algorithmic}
\end{algorithm}

\begin{figure}[t]
    \centering
    \includegraphics[width=1\columnwidth]{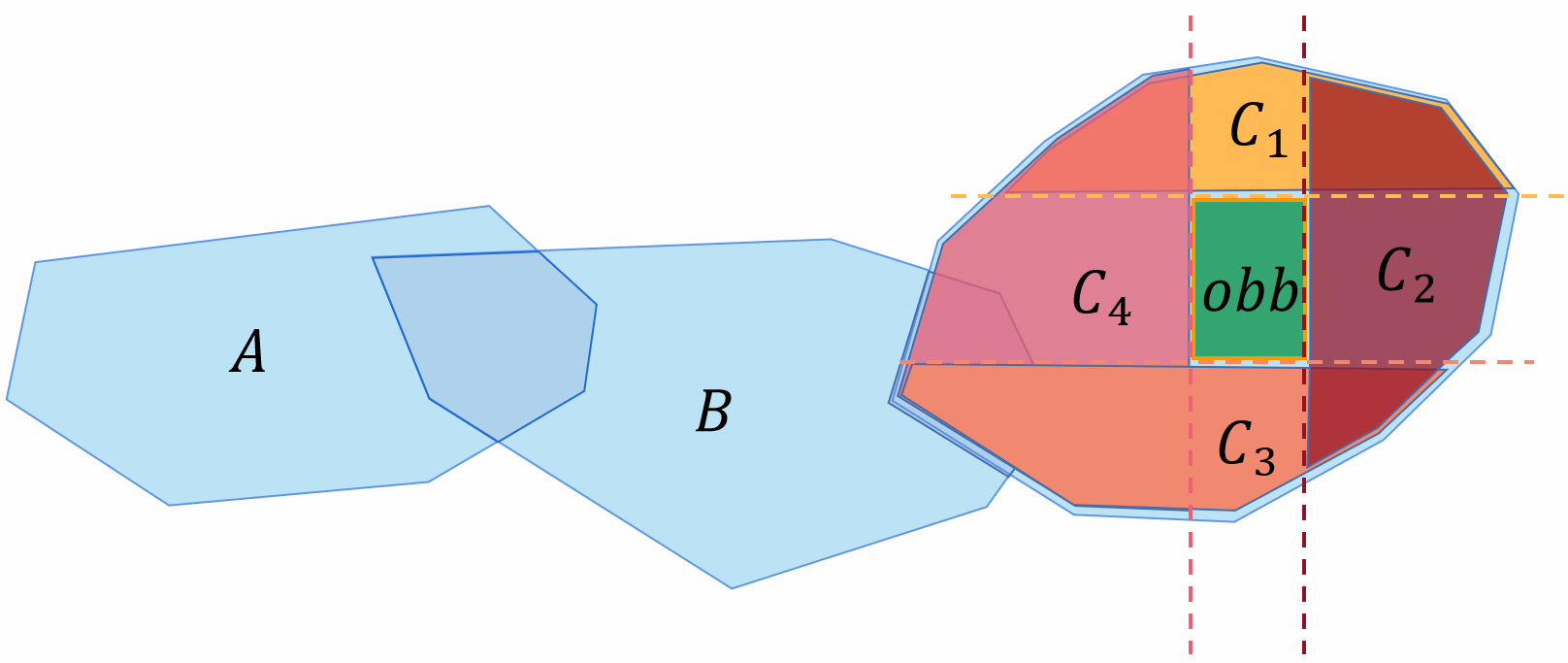}
    \caption{Illustration of avoiding unnecessary connectivity checks by prior knowledge.
    }
    \label{pic:checks}
    \vspace{-1.3cm}
\end{figure}

    Corresponding to polyhedron decomposition, polyhedron restoration is also a necessary part. Any polyhedron $\mathcal{P}_i$ decomposed by the dynamic obstacle $O_i$ should be restored as before. For the situation that the obstacle $O_i$ moves to another place, we first restored all elements in $\mathcal{P}_i$ to their original shape. Afterward, polyhedra will be decomposed by obstacles that still intersecting them. Algorithm \ref{algorithm:polyrestoration} describes the process in detail.

    \subsubsection{Graph Construct and Update}

    Connectivity between each polyhedron can be naturally used for path searching. A global graph can be constructed base on it for search-based planning methods. We interpret the geometric center of the polyhedron as room and the geometric center of the intersection between polyhedra as doors. The rooms are vertices of the graph thus going from one room to another adjacent room will go through doors. Thanks to convex set property, edges that connect any two vertices in graph are collision free. The graph will be updated after every polyhedron decomposition and restoration. In order to update the graph efficiently, we take full advantage of the prior information between the convex hulls to avoid unnecessary connectivity checks. As illustrated in Fig. \ref{pic:checks}, there is no need to check the connectivity between Polyhedron $C_1$ and $C_3$, $C_2$ and $C_4$ since they are naturally disjoint. The connectivity between $C_1-C_4$ can be established quickly. Moreover, we do not need to check the connectivity between polyhedron $A$ and each of $C_1-C_4$ by utilizing the prior information of the graph.

	\section{Experiments and Results}
	\label{sec:results}

    \begin{figure*}[t]
    \centering
    \includegraphics[width=1.95\columnwidth]{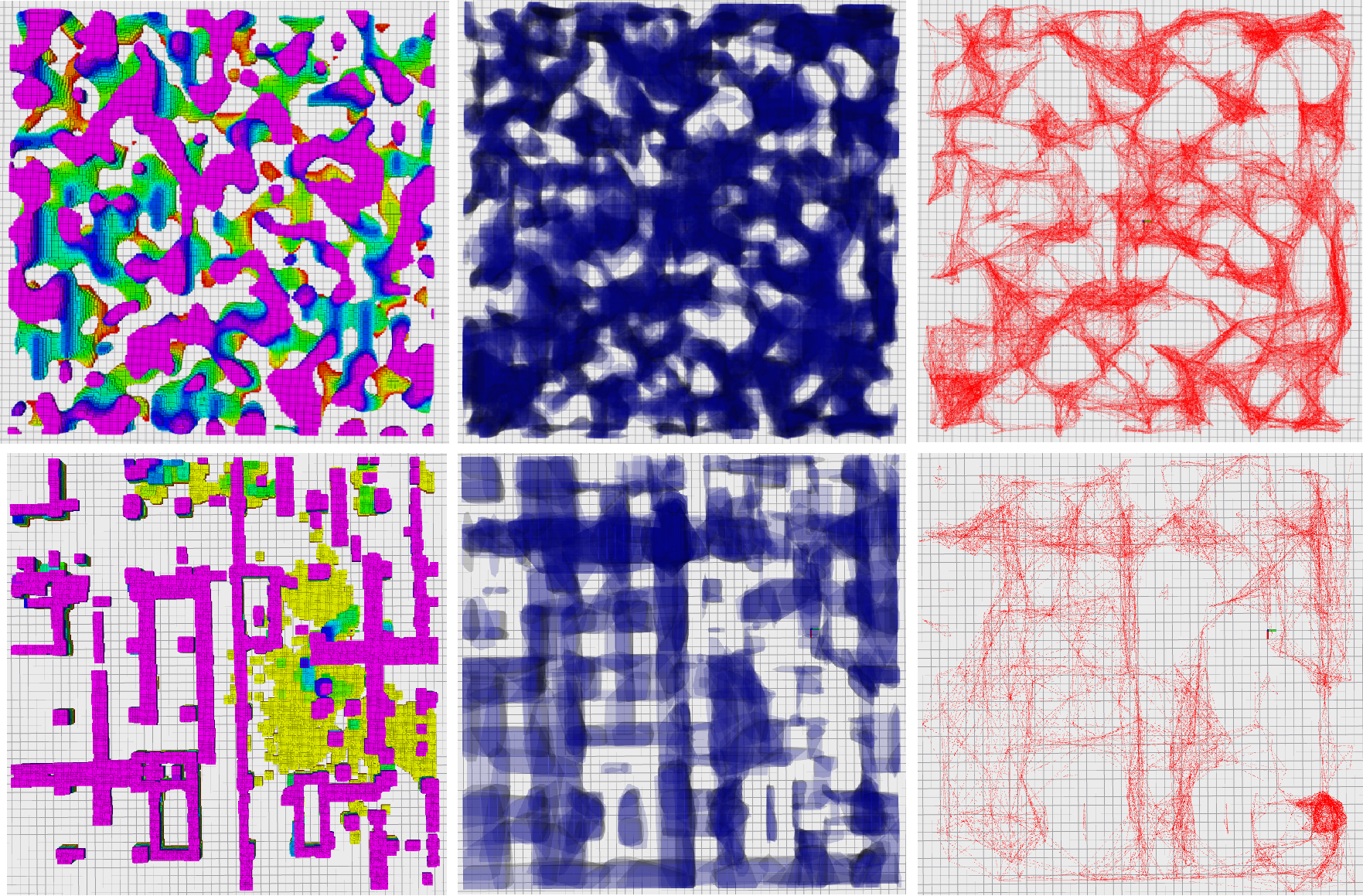}
    \caption{The detailed informantion of cluster and sparse scenario. the grid map,free-space roadmap and navigation graph of sparse scenario are presented on the top while the bottom corresponds to the sparse.
    }
    \label{pic:simu}
    \vspace{-0.2cm}
    \end{figure*}

	 To validate the applicability and evaluate the efficiency of our algorithms, simulations and real-world experiments are both conducted. For feasibility validation, we conduct motion planning in simulations and real-world experiments to demonstrate that our roadmap can provide safe feasible spaces for our planners in real-time. While for efficiency evaluation, we mainly focus on roadmap update efficiency. We evaluate the map update efficiency by computing time for polyhedron decomposition $t_d$, the computing time for polyhedron restoration $t_r$ and the computing time for graph update $t_g$.

	 \subsection{Simulation Experiments}
	 In simulation experiments, we validate our algorithm in both clustered and sparse scenarios to demonstrate that our algorithm is feasible and efficient. The size of two scenarios are set to the same. There are four dynamic obstacles in both scenarios. The detailed information of two scenarios is shown in Fig. \ref{pic:simu}. In clustered scenario, both the map and the dynamic obstacles are random generated. The detailed specifications of map are illustrated in Tab. \ref{tab:map}. While in sparse scenario, the map is constructed according to a underground garage in real world. The dynamic obstacles are also random generated. In clustered envieonment, there are totally $2613$ polyhedra generated, and $37753$ edges in the graph. While in sparse scenario, there are only $651$ polyhedra, and $5468$ edges in the graph.

	 	 \begin{table}[b]
		\centering
		\caption{Specifications of Two Maps.}
		\setlength{\tabcolsep}{2.0mm}
		\renewcommand\arraystretch{1.2}
		{
			\begin{tabular}{|c|c|c|}
				\hline
				& \tabincell{c}{Map Size} & \tabincell{c}{Resolution}\\
				\hline
				Cluster Scenario & $50m \times 50m \times 5m$ & $0.2m \times 0.2m \times 0.2m$  \\
				\hline
				Sparse Scenario &  $50m \times 50m \times 5m$  & $0.2m \times 0.2m \times 0.2m$  \\
			    \hline
		\end{tabular}}
		\label{tab:map}
	    \end{table}

\begin{figure}[t]
		\centering
		\includegraphics[width=1.0\columnwidth]{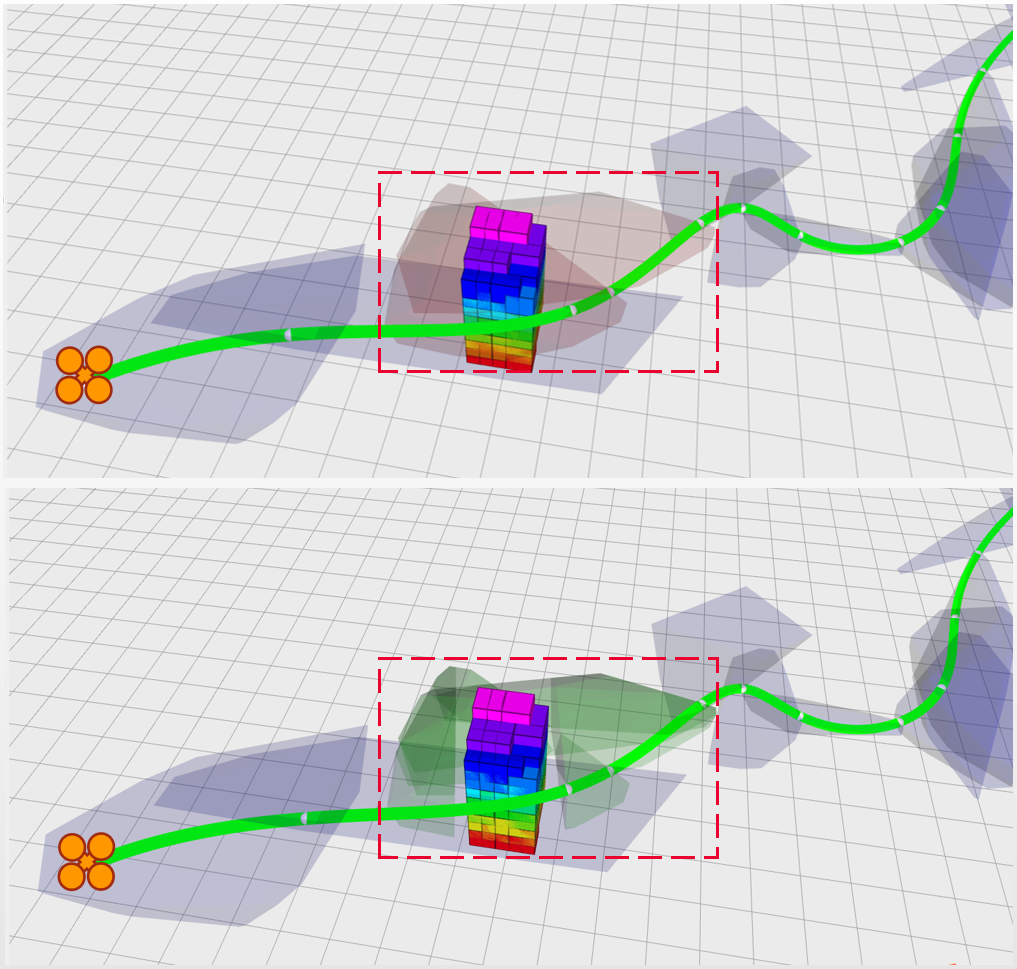}
		\caption{Feasibility validation. Once there is an obstacle occur in convex hulls (red color) we want to pass through, the convex hulls (red color) will be decomposed into some small obstacle-free convex hulls (green color).}
		\label{pic:simu2}
        \vspace{-1.0cm}
	    \end{figure}

     We conduct trajectory planning in both scenarios to validate the feasibility of our algorithm. We randomly select the start and goal point in free spaces, then $A^*$ is applied to search a series of polyhedra connecting the start and goal in the graph. Finally, A collision-free trajectory is generated by a global trajectory optimizer \cite{wang2021robust}. The re-planning frequency is set to $10Hz$ and triggered whenever the polyhedra the vehicle is going through has been decomposed or restored in last $0.2s$. Results illustrated in Fig. \ref{pic:simu2} demonstrate that when dynamic obstacles occur in the polyhedra that the quadrotor are going through, these polyhedra is decomposed efficiently and the graph is updated at the same time. After that $A^*$ finds another series of free polyhedra connecting the current point and the goal. The results validate that the roadmap is able to reflect changes in the environment for real-time safe motion planning.

    \begin{table}[t]
    \vspace{0.0cm}
    \centering
    \caption{Time Metrics of Cluster Scenario Simulation.}
    \setlength{\tabcolsep}{2.0mm}
    \renewcommand\arraystretch{1.2}
    {
        \begin{tabular}{|c|c|c|c|}
            \hline
            & \tabincell{c}{Times} & \tabincell{c}{Total Time} &Average Time\\
            \hline
            Decomposition  &  853  &  $2.375s$ & $2.784ms$ \\
            \hline
            Restoration &  $809$  & $0.390s$ & $0.482ms$   \\
            \hline
            Graph Update & $801$ & $20.039s$ & $25.017ms$ \\
            \hline

        \end{tabular}
    }
    \label{tab:cluster}
    \vspace{0.5cm}
\end{table}

\begin{table}[t]
    \vspace{0.0cm}
    \centering
    \caption{Time Metrics of Sparse Scenario Simulation.}
    \setlength{\tabcolsep}{2.0mm}
    \renewcommand\arraystretch{1.2}
    {
        \begin{tabular}{|c|c|c|c|}
            \hline
            & \tabincell{c}{Times} & \tabincell{c}{Total Time} &Average Time\\
            \hline
            Decomposition  &  $1140$  &  $3.166s$ & $2.777ms$ \\
            \hline
            Restoration &  $1527$  & $0.55458s$ & $0.3632ms$   \\
            \hline
            Graph Update & $1115$ & $18.763s$ & $16.823ms$ \\
            \hline

        \end{tabular}
    }
    \label{tab:sparse}
    \vspace{-0.5cm}
\end{table}

     We also evaluate the efficiency of our algorithm by the three metrics in both scenarios. Results are shown in both Tab. \ref{tab:cluster} and Tab. \ref{tab:sparse}. In clustered scenario, the average time of polyhedron decomposition and restoration are $2.784ms$ and $0.482$, and the average time of graph update is $25.017ms$. While in sparse scenario, the average time of polyhedron decomposition and restoration are $2.777ms$ and $0.3632ms$, and the average time of graph update is $16.823ms$. Comparing results in these two scenarios, we can see that the average time of $t_d$ and $t_r$ have negligible difference in two scenarios. However the average time of graph update $t_g$ varies a lot. This is reasonable since it requires more convex hulls to approximate the environment in clustered scenario than a sparse one in same size space. In a space with fixed size, the greater the number of convex hulls is, the more complex the connections between the convex hulls become, thus resulting in a slower graph update efficiency. That is why we utilize the prior information of the graph to avoid unnecessary connectivity checks. Even the update time of the map varies in this two scenarios, it still meets the real-time requirements.
     	 \begin{figure}[t]
		\centering
		\includegraphics[width=1.0\columnwidth]{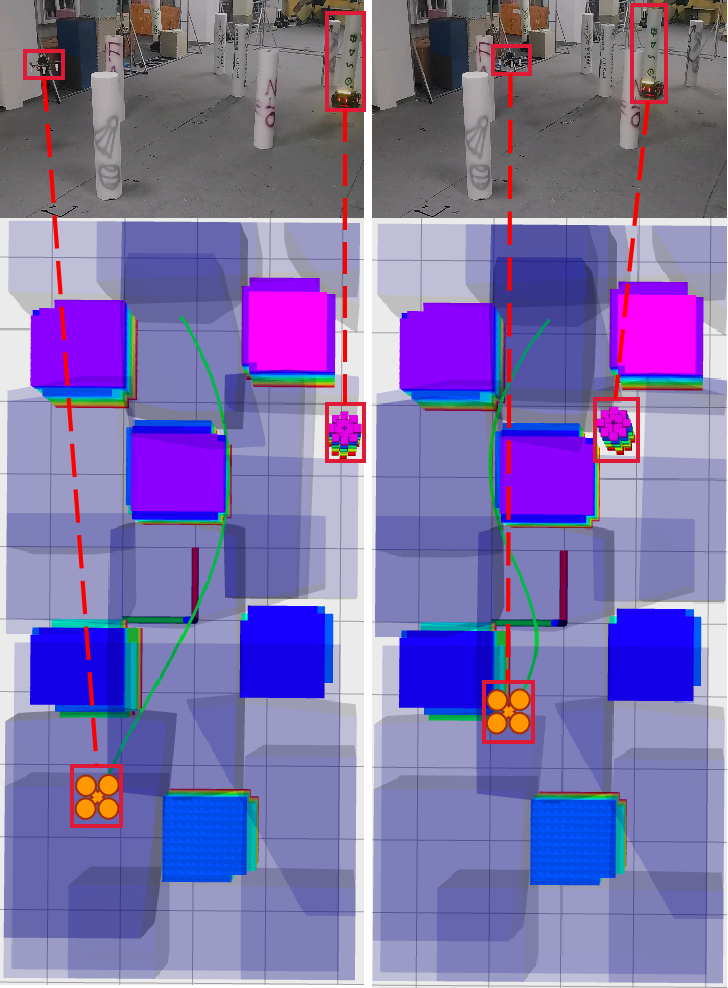}
		\caption{Left figures show that A star find a path that through the right side of the middle pillar. As the obstacle moves, the trajectory was blocked, re-plann mechaand the algorithm finds another collision-free path.}
		\label{pic:exp2}
        \vspace{-0.2cm}
	\end{figure}

	\subsection{Real-World Experiments}
	For real-world experiments, we integrate our roadmap into a quadrotor platform and use FAST-LIO \cite{xu2021fast} to pre-build the environment. We use the combination of foam pillars and a ground vehicle controlled by human operators as dynamic obstacles. The location of the quadrotor and dynamic obstacles is provided by motion capture. The quadrotor platform and the obstacles are shown in Fig.~\ref{pic:exp2}. The experiments was divided into two parts, that are to validate the feasibility and evaluate the efficiency of our algorithm. In the first part, the start point is fixed, then we manually select a goal point. The quadrotor searches a series of collision-free polyhedra connecting the start point and goal point, then accomplishes the flight. The Max velocity of the quadrotor is set to be $1m/s$. In the second part, we mainly focus on evaluate the efficiency, so there are only the obstacle move freely in the environment and the Max velocity is $6.5m/s$.
		    \begin{table}[h]
		\vspace{0.0cm}
		\centering
		\caption{Time Metrics of Real-World Experiments.}
		\setlength{\tabcolsep}{2.0mm}
		\renewcommand\arraystretch{1.2}
		{
			\begin{tabular}{|c|c|c|c|}
				\hline
				& \tabincell{c}{Times} & \tabincell{c}{Total Time} &Average Time\\
				\hline
				 Decomposition  &  $1070$  &  $1.6806s$ & $1.569ms$ \\
				 \hline
				 Restoration &  $1069$  & $0.151254s$ & $0.1414ms$   \\
				\hline
				 Graph Update & $1070$ & $3.735s$ & $3.491ms$ \\
				 \hline
		    \end{tabular}
		}
		\label{tab:real}
		\vspace{0.0cm}
	    \end{table}

    In the first experiments, we set the start point and goal point as the two ends of the map. We try to manipulate the obstacle to block the flight of quadrotor. Since the roadmap is efficiently updated and the connectivity is preserved, the quadrotor quickly finds a series of polyhedra connecting start and goal when obstacle moves. Finally the quadrotor successfully avoid the dynamic obstacle and reach the goal point as shown in Fig. \ref{pic:exp2}.
    The results of second experiments are shown in Tab. \ref{tab:real}, these three metrics reflect that the roadmap is updated efficiently. Comparing to experiments in simulations, $t_d$, $t_r$, $t_g$ in real-world environment is much lower because the map is smaller and sparser. Meanwhile, there are only one dynamic obstacle, hence there will only be two levels in the hierarchy at most.

	\section{Conclusion}
	\label{sec:conclusion}

	In this paper, we present dynamic free-space roadmap, a novel map representation that provides both safe feasible space and navigation graph. The roadmap can be incrementally updated by using our polyhedron decomposition and restoration method. We adopt the hierarchy structure to manage the map, making our map convenient to maintain. By utilizing the prior information of the graph, unnecessary connectivity checks are avoided thus the graph can be updated efficiently. We also conduct simulations and real-world experiments to demonstrate that our method is feasible and efficient. In the future, we will design a path search method and trajectory planner that fit our roadmap. A prediction module for dynamic obstacles will also be incorporated into our quadrotor to achieve autonomous flights in dynamic environments.

\newlength{\bibitemsep}\setlength{\bibitemsep}{0.20\baselineskip}
\newlength{\bibparskip}\setlength{\bibparskip}{0pt}
\let\oldthebibliography\thebibliography
\renewcommand\thebibliography[1]{
    \oldthebibliography{#1}
    \setlength{\parskip}{\bibitemsep}
    \setlength{\itemsep}{\bibparskip}
}

	\bibliography{Dynamic_Free-Space_Roadmap_for_Safe_Quadrotor_Motion_Planning}
\end{document}